\documentclass[smallcondensed]{svjour3}
\usepackage[colorlinks=true, urlcolor=black, linkcolor=black, citecolor = black]{hyperref}

\include{packages} 

\usepackage{xr}
\usepackage{algorithm}
\usepackage{algorithmic}
\usepackage[round]{natbib}
\usepackage{lmodern}

\usepackage{microtype}
\usepackage{graphicx}
\usepackage{subfigure}
\usepackage{booktabs} 
\usepackage{multirow}

\titlerunning{Joint Detection of Malicious Domains and Infected Clients}

\title{Joint Detection of Malicious Domains and Infected Clients\footnote{This is a pre-print of an article published in Machine Learning. The
final version is available online at https://doi.org/10.1007/s10994-019-05789-z}}
\author{Paul Prasse$^1$ \and Ren\'e Knaebel$^1$ \and Luk\'{a}\v{s} Machlica$^2$ \and Tom\'{a}\v{s} Pevn\'{y}$^{2, 3}$ \and Tobias Scheffer$^1$}
\authorrunning{P. Prasse, R. Knaebel, L. Machlika, T. Pevn\'{y}, T. Scheffer}
\institute{$^1$University of Potsdam, Department of Computer Science, Potsdam, Germany, \{prasse, knaebel, scheffer\}@cs.uni-potsdam.de \and
$^2$Cisco R\&D, Prague, Czech Republic, \{lumachli, tpevny\}@cisco.com \and $^3$Czech Technical University in Prague, Department of Computer Science
}

\journalname{This is a pre-print of an article published in Machine Learning.}

\begin{document} 

\maketitle
\vskip 0.3in




%

\newcommand\eg{{\em e.g.,}}

\begin{abstract}
Detection of malware-infected computers and detection of malicious web domains based on their encrypted HTTPS traffic are challenging problems, because only addresses, timestamps, and data volumes are observable. The detection problems are coupled, because infected clients tend to interact with malicious domains. Traffic data can be collected at a large scale, and antivirus tools can be used to identify infected clients in retrospect. Domains, by contrast, have to be labeled individually after forensic analysis. We explore transfer learning based on sluice networks; this allows the detection models to bootstrap each other. In a large-scale experimental study, we find that the model outperforms known reference models and detects previously unknown malware, previously unknown malware families, and previously unknown malicious domains.
\end{abstract}


%

\section{Introduction}

Malware violates users' privacy, harvests passwords and personal information, can encrypt users' files for ransom, is used to commit click-fraud, and to promote political agendas by popularizing specific content in social media~\citep{Kogan2015bedep}. Client-based antivirus tools use vendor-specific blends of signature-based analysis, static analysis of portable-executable files, emulation (partial execution without access to actual system resources prior to execution in the actual operating system) and dynamic, behavior-based analysis to detect malware~\citep{swinnen2014one}. Network-traffic analysis complements antivirus software and is widely used in corporate networks. Traffic analysis allows organizations to enforce acceptable-use and security policies consistently throughout the network and minimize management overhead. Traffic analysis makes it possible to encapsulate malware detection into network devices or cloud services that can detect polymorphic malware~\citep{karim2005malware} as well as yet-unknown malware based on, for instance, URL patterns~\citep{bartos2015robust}.

However, malware can easily prevent the analysis of its HTTP payload by using the encrypted \textit{HTTPS} protocol. 
The use of HTTPS by itself is not conspicuous because Google, Facebook, LinkedIn, and many other popular sites encrypt their network traffic by default and the global data volume of HTTPS has surpassed that of HTTP~\citep{wired2017half}. In order to subject HTTPS traffic to network-traffic analysis, organizations today have to configure their network such that all web traffic is routed via a web-security server. This server's root certificate has to be installed as a trusted certificate on all client computers, which allows the service to act as a man-in-the-middle between client and host. It can decrypt, inspect, and re-encrypt HTTPS requests. This approach scales poorly to large networks because the cryptographic operations are computationally expensive, and it introduces a potential vulnerability into the network. 

Without breaking the encryption, an observer of HTTPS traffic can only see the client and host IP addresses and ports, and the timestamps and data volumes of packets. Network devices aggregate TCP/IP packets exchanged between a pair of IP addresses and ports into a {\em network flow} for which address, timing, and data-volume information are saved to a log file. Most of the time, an observer can also see the unencrypted host domain name.
The HTTP payload, including the HTTP header fields and the URL, are encrypted. 

Web hosts are involved in a wide range of illegitimate activities, and blacklisting traffic to and from known malicious domains and IP addresses is an effective mechanism against malware. Malicious domains can host back-ends for banking trojans and financial scams, click-fraud servers, or distribution hubs for malicious content. Identifying a domain as malicious requires a complex forensic analysis. An analyst has to collect information about the server that hosts the domain, software and employed technologies, and can research ownership of the domain and co-hosted domains as well as observe the host's behavior. 

Since many types of malicious activities involve interaction with client-based malware, the detection of malicious hosts and infected clients are coupled problems. In the context of neural networks, labeled data for related tasks are often exploited by designing coupled networks that share part of the parameters. In sluice networks~\citep{ruder2017sluice}, the extent to which parameters are shared is itself controlled by parameters, which allows auxiliary data to serve as a flexible prior for the task at hand. 

The rest of this paper is structured as follows. Section~\ref{sec:rel} reviews related work. We describe our operating environment and our data in Section~\ref{sec:arch} and the problem setting in Section \ref{sec:ClientMalwareDetectionProblem}. In Section \ref{sec:detection}, we derive a model for joint detection for malware and malicious domains and describe reference methods. Section~\ref{sec:experiments} presents experiments; Section~\ref{sec:conclusion} concludes.

\section{Related Work}\label{sec:rel}

Prior work on the analysis of {\em HTTP logs}~\citep{nguyen2008survey} has addressed the problems of identifying command-and-control servers~\citep{nelms2013execscent}, unsupervised detection of malware~\citep{kohout2015unsupervised,bartos2016optimized}, and supervised detection of malware using domain blacklists as labels~\citep{franc2015learning,bartos2015robust}.
HTTP log files contain the full URL string, from which a wide array of informative features can be extracted~\citep{bartos2015robust}. 

A body of recent work has aimed at detecting Android malware by network-traffic analysis. \citet{arora2014malware} use the average packet size, average flow duration, and a small set of other features to identify a small set of 48 malicious Android apps with some accuracy. \citet{lashkari2015towards} collect 1,500 benign and 400 malicious Android apps, extract flow duration and volume feature, and apply several several machine-learning algorithms from the Weka library. They observe high accuracy values on the level of individual flows. 
\citet{demontis2017yes} model different types of attacks against such detection mechanisms and devise a feature-learning paradigm that mitigates these attacks.
\citet{malik2016credroid} aggregate the VirusTotal ranking of an app with a crowd-sourced domain-reputation service (Web of Trust) and the app's resource permission to arrive at a ranking. 

Prior work on {\em HTTPS logs} has aimed at identifying the application layer protocol~\citep{wright2006on,crotti2007traffic,dusi2009tunnel}. In order to cluster web servers that host similar applications,~\citet{kohout2015automatic} develop features that are derived from a histogram of observable time intervals and data volumes of connections. Using this feature representation,~\citet{lokovc2016k} develop an approximate $k$-NN classifier that identifies servers which are contacted by malware. Hosts that are contacted by malware are by no means necessarily malicious. Malware uses URL forwarding and other techniques to route its traffic via legitimate hosts, and may contact legitimate services just to dilute its network traffic. We will nevertheless use the histogram features as a reference feature representation.

Graph-based classification methods \citep[\eg][]{anderson2011graph} have been explored but cannot be applied in our operating environment. In our operating environment, a Cloud Web Security server observes only the network traffic within an organization. In order to perceive a significant portion of the network graph, companies would have to exchange their network-traffic data which is impractical for logistic and privacy reasons. 

Prior work on neural networks for network-flow analysis~\citep{pevny2016discriminative} has worked with labels for client computers (infected and not infected)---which leads to a multi-instance learning problem. By contrast, our operating environment allows us to observe the association between flows and executable files. Malware detection from HTTPS traffic has been studied using a combination of word2vec embeddings of domain names and long short term memory networks (LSTMs)~\citep{prasse2017ecml}. We will use this method as a reference in our experiments. Recent findings suggest that the greater robustness of convolutional neural networks (CNNs) outweights the ability of LSTMs to account for long-term dependencies~\citep{gehring2017convolutional}. This motivates us to explore convolutional architectures. Neural networks have also been applied to static malware analysis~\citep{pascanu2015malware}.

In the context of deep learning, multi-task learning is most often implemented via hard or soft parameter sharing of hidden layers. In hard parameter sharing, models for all task can share the convolutional layers~\citep{long2017learning} or even all hidden layers~\citep{caruana1993multitask}, which can dramatically increase the sample size used to optimize most of the parameters~\citep{baxter1997bayesian}. Soft parameter sharing, by contrast, can be realized as a direct application of hierarchical Bayesian modeling to neural network: each parameter is regularized towards its mean value across all tasks~\citep{duong2015neural,yang2016trace}. Cross-stitch~\citep{misra2016cross}  and sluice networks~\citep{ruder2017sluice} allow the extent of task coupling for separate parts of the network to be controlled by parameters. Sluice networks have a slightly more general form than cross-stitch networks because they have additional parameters that allow a task-specific weighting of network layers. 

Alternative transfer-learning approaches for neural networks enforce an intermediate representation that is invariant across tasks~\citep{ganin2016domain}. Outside of deep learning, the group lasso regularizer enforces  subspace sharing, and wide range of approaches to multi-task learning have been studied, based on hierarchical Bayesian models (\eg {} \citealp{finkel2009hierarchical}), learning task-invariant features (\eg{} \citealp{argyriou2007multi}), task-similarity kernels~\citep{evgeniou2005learning}, and learning instance-specific weights (\eg{} \citealp{bickel2008multi}).

\section{Operating Environment}\label{sec:arch}

This section describes our application environment.
In order to protect all computers of an organization, a Cloud Web Security (CWS) service provides an interface between the organization's private network and the internet. Client computers establish a VPN connection to the CWS service, and all external HTTP and HTTPS connections from any client within the organization is then routed via this service. The service can block HTTP and HTTPS requests based on the host domain and on the organization's acceptable-use policy. The CWS service blocks all traffic to and from all malicious domains on a curated blacklist. It issues warnings when it has detected malware on a client. Since security analysts have to process the malware warnings, the proportion of false alarms among all issued warnings has to be small. 

On the application layer, HTTPS uses the HTTP protocol, but all messages are encrypted via the Transport Layer Security (TLS) protocol or its predecessor, the Secure Sockets Layer (SSL) protocol. 
The CWS service aggregates all TCP/IP packets between a single client computer, client port, host IP address, and host port that result from a single HTTP request or from the TLS/SSL tunnel of an HTTPS request into a {\em network flow}. 
For each network flow, a line is written into the log file that includes data volume, timestamp, client and host address, and duration information. For unencrypted HTTP traffic, this line also contains the full URL string. For HTTPS traffic, it includes the domain name---if that name can be observed via one of the following mechanisms. 

Clients that use the Server Name Indication protocol extension (SNI) publish the unencrypted host-domain name when they establish the connection. SNI is widely used because it is necessary to verify certificates of servers that host multiple domains, as most web servers do. When the network uses a transparent DNS proxy~\citep{blum2001transparent}, this server caches DNS request-response pairs and can map IP addresses to previously resolved domain names. 
The resulting sequence of log-file lines serves as input to the detection models for malware and malicious domains.

\subsection{Data Collection}\label{sec:data}

For our experiments, we combine a large collection of HTTPS network flows~\citep{prasse2017ecml} that have been labeled by whether they originate from a malicious or legitimate application with a domain blacklist that is maintained by forensics experts at Cisco.

\citet{prasse2017ecml} have collected the HTTPS network flows that pass CWS servers in 340 corporate networks. The client computers in these networks run a VPN client that monitors the process table and network interface, and keeps a record of which executable file creates each network flow. In retrospect, the executable files have been analyzed with a multitude of antivirus tools. The resulting data set consists of network flows between known clients (identified by organization and VPN account), domains (fully qualified domain names), data volumes and timestamps, and a label that indicates whether the application that generated the traffic is recognized as malware by antivirus tools.
We stratify training and test data in chronological order. The {\em training data} contains the complete HTTPS traffic of 171 small to large computer networks for a period of 5 days in July 2016. 
The \textit{test data } contains the complete HTTPS traffic of 169 different computer networks for a period of 8 days in September 2016. 
Forensics experts at Cisco continuously investigate suspicious host names, second-level domain names, and server IP addresses that have been flagged by a wide range of mechanisms. This includes an analysis of the hosted software and employed technologies, of registry records, URL and traffic patterns, and any additional information that may be available for a particular domain. We believe that domains are almost never erroneously rated as malicious, but due to the expensive analytic process, the blacklist of malicious domains is necessarily incomplete. 
All traffic from and to malicious serves can easily be blocked by the CWS service. The network traffic does not contain any flows to domains that had been on our blacklist at the time when the traffic data were collected. The traffic data set contains network flows to and from 4,340 malicious host names, second-level domains, and server IP addresses that have been added to the blacklist after the data were collected.

\subsection{Quantitative Analysis of the Data}

Table~\ref{tab:numdata} and Table~\ref{tab:numapps} summarizes the number of benign and malicious network flows, client computers, infected computers, applications with unique hashes, and organizations. 

Table \ref{tab:malwareTypes} gives statistics about the most frequent malware families.
It enumerates the number of variations that occur, the number of infected clients, and, in parentheses, the number of infected clients in the training data.

In total, just below 18,000 computers are malware-infected and communicate with domains that had not been blacklisted at the time, which corresponds to almost 0.6\%.

In the traffic data, 4,340 domains occur that have been added to the blacklist after the traffic data were recorded. Table~\ref{tab:serverstats} details the types of malicious host names, second-level domains, and server IP addresses that occur in all data and in the training data.

\begin{table*}[th]
\caption{Key statistics of the HTTPS network-traffic data sets.}\label{tab:numdata}
\begin{center}
\begin{tabular}{l||r|r|r||r|r||r}
data set & flows & malicious & benign & users & infected & organizations
\\ \hline
training& 44,348,879 & 350,220 & 43,150,605 & 133,437 & 8,944  & 171\\ 
test & 149,005,149 & 955,037 & 142,592,850 & 177,738 &8,971 & 169\\ 
\end{tabular}
\end{center}
\end{table*}

\begin{table*}[th]
\caption{Number of applications in HTTPS network-traffic data sets.}\label{tab:numapps}
\begin{center}
\begin{tabular}{l||r|r}
data set & applications & malicious
\\ \hline
training&  20,169 & 1,168\\ 
test &  27,264 & 1,237\\ 
\end{tabular}
\end{center}
\end{table*}

\begin{table}[t!]
\caption{Malware families and malware types.}\label{tab:malwareTypes}
\center
\begin{tabular}{l|r|r}
{malware family} & {variations} & clients\\ \hline
dealply& 506			 & 1,385 (516)			\\ 
softcnapp &119		 & 797 (250)			\\ 
crossrider& 98		 & 274 (102)			\\ 
elex&86					 & 779 (316)			\\ 
opencandy &57 & 164 (126)				\\ 
conduit&56				 & 314 (103)	\\
browsefox&52			 & 78 (34)		\\ 
speedingupmypc&29 & 224 (63)			\\
kraddare& 28	     & 33 (26)						\\ 
installcore& 27	 & 49 (19)		\\ 
mobogenie&26			 & 467 (184)		\\ 
pullupdate&25		 & 99 (25)			\\ 
iobit downloader & 24 & 38 (15)							\\ 
asparnet &24	 & 5,267 (5,128)			\\ 
\end{tabular}
\end{table}


\begin{table}
\caption{Domain-label statistics}\label{tab:serverstats}
\begin{center}
\begin{tabular}{l|r|r}
type & total & training\\
\hline
	  malware-distribution& 2730 &478\\
		ad-injector& 961 &576\\
		malicious-content-distribution& 276 & 171\\
		potentially unwanted application& 97 & 75\\
		click-fraud& 65 & 50\\
		spam-tracking & 61 & 51 \\ 
		information-stealer& 52& 22\\
		scareware& 30 & 23\\
		money-scam& 22 & 9\\
		banking-trojan& 19 & 10\\
    malicious-advertising& 13 & 10\\
    cryptocurrency-miner& 9 & 0\\   
		ransomware& 3 &  3\\
		anonymization-software& 2 & 1\\
\end{tabular}
\end{center}
\end{table}

\section{Problem Setting}
\label{sec:ClientMalwareDetectionProblem}

We will now establish the problem setting.
Our goal is to flag client computers that are hosting malware, and to flag malicious web domains. Client computers are identified by a (local) IP address and a VPN user name; web domains are identified by a fully qualified domain name or, when no domain name can be observed, an IP address. 

We have two types of classification instances. For each interval of 24 hours, we count every client computer that establishes at least one network connection as a separate \textit{classification instance} of the malware-detection problem. A client that is active on multiple days constitutes multiple classification instances; this allows us to issue daily infection warnings for clients. Equivalently, for each interval of 24 hours, we model each observed fully qualified domain name as a classification instance. This allows us to make daily blacklisting decisions, and to disregard traffic after 24 hours in the deployed system. 
    
Our training data are labeled at the granularity of a network flow between a client and a host.
This allows us to train classification models at the granularity of network flows. Learning a network-flows classifier from labeled flows is an intrinsically easier problem than learning a detection model from labels at the granularity level of clients or domains. While a detection model that is trained from labeled clients or domains has to figure out which flows pertain to the malicious activity, the network-flow classification model is handed that information during training. 

Since a client is infected if it is running at least one malicious application and a domain is malicious if it engages in at least one malicious activity, it is natural to aggregate the classification results for network flows into detection results for clients and domains by max-pooling the decision-function values over all flows for that client or domain, respectively, throughout the period of 24 hours. 
The flow classifiers are thereby applied as one-dimensional convolutions over time; max-pooling the outcome yields detection models for infected clients and malicious domains. 
Since an application generally generates multiple network flows, it may be helpful to take the context into account when classifying each  flow. Our input representation therefore includes a window of the client's flows that is centered over the flow to be classified. The width of this window is a model parameter. This window always contains the context of network flows for a client, both for detection of malware and of malicious domains. While the CWS server can observe the complete traffic of each client in the network, it will generally only observe a small fraction of traffic to and from a domain outside the network.

We will measure precision-recall curves because they are most directly linked to the merit of a detection method from an application point of view. Precision---the fraction of alarms that are not false alarms---is directly linked to unnecessary workload imposed on security analysts, while recall quantifies the detection rate.
However, since precision-recall curves are not invariant in the class ratio, we will additionally use ROC curves to compare the performance of classifiers on data sets with varying class ratios. 
Note that the relationship between false-positive rate and precision depends on the class ratio. 
For instance, at a false-positive rate of 10\%, the expected number of false alarms equals 10\% of the number of benign instances; hence, false alarms would by far outnumber actual detections. By contrast, at a precision of 90\%, only 10\% of all alarms would be false alarms.

\section{Network-Flow Analysis}\label{sec:detection}

This section presents our architecture that jointly detects infected clients and malicious domains, as well as reference models that we will compare against.

\subsection{Sluice Network}

\begin{figure}[t!]
\centerline{\includegraphics[width=0.75 \columnwidth]{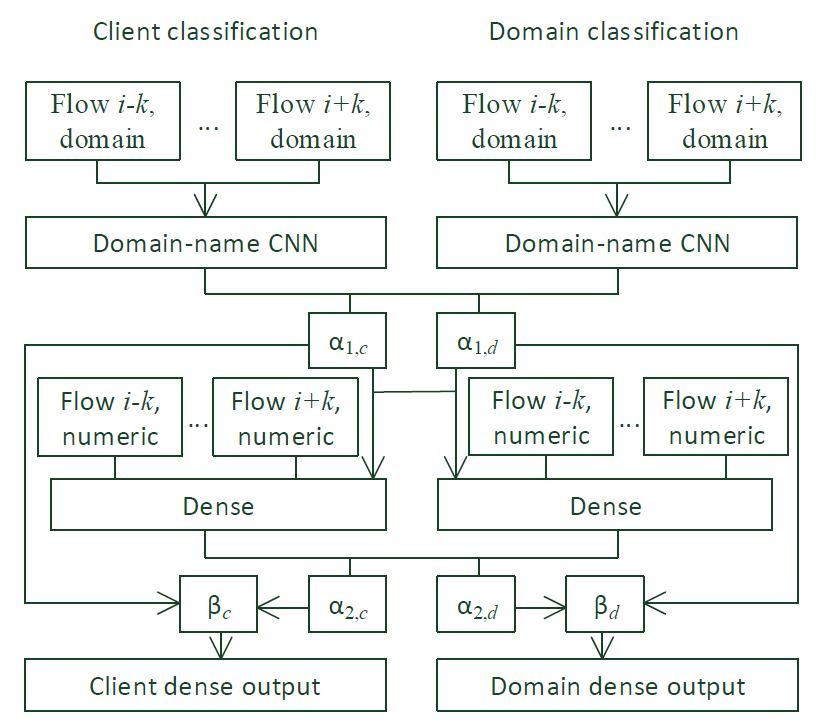}}
\caption{Sluice Dense on Domain CNN}\label{fig:sluiceDenseonCNN}
\end{figure}

\begin{figure}[t!]
\centerline{\includegraphics[width=0.99 \columnwidth]{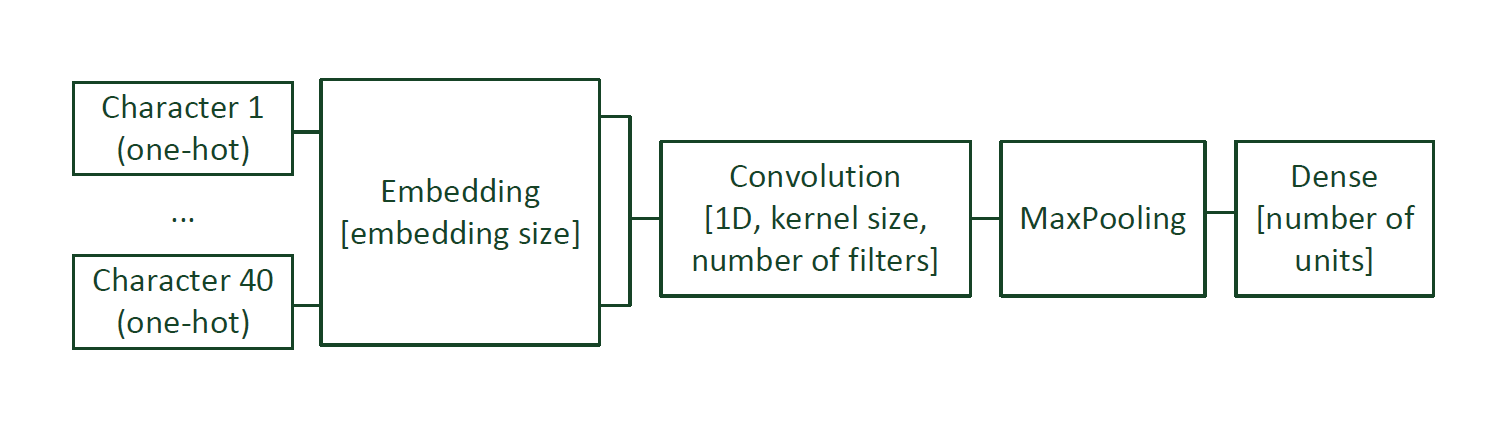}}
\caption{Domain CNN architecture}\label{fig:domainCNN}
\end{figure}

Figure~\ref{fig:sluiceDenseonCNN} shows the \textit{sluice} network architecture for joint flow classification with soft parameter sharing. The left-hand part of the network classifies flows by whether they originate from infected clients, the right-hand part classifies flows by whether they are between a client and a malicious domain. The input features are derived from a window of $2k+1$ flows for a given client that is centered around the flow to be classified. The first stage of the network---the domain-name CNNs---receives the domain names of the host domains within that window as input. 

Figure~\ref{fig:domainCNN} shows this domain-name CNN in more detail. It has a standard convolutional architecture with convolutional, max-pooling, and dense layers. 
Domain names are first represented as one-hot-encoded character sequences of the up to 40 last characters of a domain name. We selected the value of 40 because further increasing this parameter does not change the experimental results. 
In the next step, an embedding layer reduces this dimensionality; weights are shared for the embedding of each character. This is followed by a one-dimensional convolutional layer, a max-pooling layer, and a dense layer that constitutes the final encoding of the domain name. 

The following dense layers receive the window of $2k+1$ domain-name embeddings. Additionally, they receive a vector of numeric features for each of the $2k+1$ flows in the input window. The numeric attributes consist of the log-transformed duration, log-transformed numbers of sent and received bytes, duration, and the time gap from the preceding flow. These dense layers are followed by softmax output layers. 

After each stage, the output from either side of the network is combined into a weighted average controlled by coupling coefficients $\alpha$. Values of $\alpha_{\cdot,\cdot}=0$ correspond to independent networks. In addition, the output layer is allowed to draw on all intermediate layers. The output of each hidden layer is weighted by a coefficient $\beta$ and all weighted outputs are concatenated. Setting all the $\beta_{\cdot}$ values associated with the first hidden layer to zero and all values associated with the second hidden layer to one correspond to the standard layered feed-forward architecture. We use the ReLU activation function for hidden layers.

The model is trained by using backpropagation on labeled network flows. At application time,  detection results at the level of clients and domains are derived by maximizing the output scores  of the positive class ``infected client'' over all network flows between the given client and any domain over an interval of 24 hours. A client is flagged as soon at this maximum exceeds a threshold value. Likewise, the output scores of the positive class ``malicious domain'' on the right-hand side is maximized over all flows between any client and the domain to be classified for 24 hours. 

\subsection{Independent Models and Hard Sharing}

Separating the left- and right-hand side of the sluice architecture constitutes the first natural baseline. We will refer to these models as {\sl independent} models. 
This is equivalent to setting $\alpha_{\cdot,\cdot}=0$, setting all the $\beta_{\cdot}$ values associated
with the first hidden layer to zero and all values associated with the second hidden layer to one. The next natural baseline is {\sl hard parameter sharing}. Here, only the output layers of the client- and domain-classification models have independent parameters while the domain CNN and the following dense layer exist only once.

\subsection{LSTM on Word2vec}

This baseline model~\citep{prasse2017ecml} uses the word2vec continuous bag-of-words model~\citep{mikolov2013distributed} to embed domain names, and processes the flow sequence with an LSTM. The input to the network consists of character $n$-grams that are one-hot coded as a binary vector in which each dimension represents an $n$-gram. The input layer is fully connected to a hidden layer that implements the embedding. The same weight matrix is applied to all input character $n$-grams. The activation of the hidden units is the vector-space representation of the input $n$-gram of characters. In order to infer the vector-space representation of an entire domain-name, an ``averaging layer'' averages the hidden-unit activations of all its character $n$-grams. 

We use the weight matrix and configuration of~\citet{prasse2017ecml} and refer to this model as {\sl LSTM on word2vec}. This model uses character $2$-grams, resulting in 1,583 character 2-grams. \citet{prasse2017ecml} have found the {\sl LSTM on word2vec} model to outperform a random-forest model. We therefore consider {\sl LSTM on word2vec} to be our reference and do not include random forests in our experiments. 

\subsection{Metric Space Learning}

\citet{lokovc2016k} extract a vector of soft histogram features for the flows between any client and a given domain. They apply a $k$-NN classifier in order to identify domains that are contacted by malware. We apply this approach to our problem of detecting malicious domains. We use exact inference instead of the approximate inference proposed by \citet{lokovc2016k}. We prop this baseline up by additionally providing a list of engineered domain features described by~\citet{franc2015learning} as input to the classifier; we refer to this method as {\sl 4-NN soft histograms}.

\section{Experiments}\label{sec:experiments}

This section reports on malware-detection and malicious-domain-detection accuracy. We train all models on a single server with 40-core Intel(R) Xeon(R) CPU E5-2640 processor and 128 GB of memory. We train all neural networks using the Keras~\citep{chollet2015keras} and Tensorflow~\citep{tensorflow2015-whitepaper} libraries on a GeForce GTX TITAN X GPU using the NVidia CUDA platform. We implement the evaluation framework using the scikit-learn~\citep{scikit-learn} machine learning package.

\subsection{Parameter Optimization}

We optimize the hyperparameters of all networks on the training data using the hyperband algorithm~\citep{Li2016hyperband}. For the domain CNN, we vary the embedding size between $2^5$ and $2^7$, the kernel size between $2$ and $2^4$, the number of filters between 2 and $2^9$, and the number of dense units between $2^5$ and $2^9$. For the {\sl sluice network}, the {\sl independent models} and {\sl hard parameter sharing}, we vary the number of dense units between $2^5$ and $2^{11}$, and the window size between 1 and 15 flows.
Table~\ref{tab:best_parameter1} and Table~\ref{tab:best_parameter2} shows the hyperparameter values after optimization.

\begin{table*}[t!]
\begin{center}
\caption{Best hyperparameters found using hyperband for models with shared blocks.}\label{tab:best_parameter1}
{\scriptsize\begin{tabular}{cl|c||l|c}
&\multicolumn{2}{c||}{{Sluice }} & \multicolumn{2}{c}{{Hard parameter sharing}} \\ 
&{hyperparameter} &  {value} & {hyperparameter} & {value} \\ \cline{1-5}
\multirow{4}{*}{\rotatebox[origin=c]{90}{\begin{minipage}{0.5in} \begin{center}Domain CNN\end{center}\end{minipage}}} & embedding size & 128  & embedding size & 64\\ 
&kernel size & 16  & kernel size & 16 \\
& filters & 512 & filters & 512 \\ 
&dense units& 256 & dense units& 64 \\ \cline{1-5}
\multirow{3}{*}{\rotatebox[origin=c]{90}{\begin{minipage}{0.2in} \begin{center}Flow\\ class.\end{center}\end{minipage}}} & dense units & 512 & dense units & 512 \\ 
&window size & 11 & window size & 7 \\
\end{tabular}}
\end{center}
\end{table*}

\begin{table*}[t!]
\begin{center}
\caption{Best hyperparameters found using hyperband for independent models.}\label{tab:best_parameter2}
{\scriptsize\begin{tabular}{cl|c||l|c}
&\multicolumn{2}{c||}{{Independent client}} & \multicolumn{2}{c}{{Independent domain}} \\ 
&{hyperparameter} &  {value} & {hyperparameter} & {value} \\ \cline{1-5}
\multirow{4}{*}{\rotatebox[origin=c]{90}{\begin{minipage}{0.5in} \begin{center}Domain CNN\end{center}\end{minipage}}} & embedding size & 64 & embedding size & 64 \\ 
& kernel size & 16 & kernel size & 16 \\
&filters & 128 &filters & 128  \\ 
&dense units& 32 &dense units& 32  \\ \cline{1-5}
\multirow{3}{*}{\rotatebox[origin=c]{90}{\begin{minipage}{0.2in} \begin{center}Flow\\ class.\end{center}\end{minipage}}} & dense units & 512  & dense units & 512\\ 
& window size & 9 & window size & 9\\
\end{tabular}}
\end{center}
\end{table*}

\begin{figure*}[h!]
	\centering
		\subfigure[Precision-recall curves]{\includegraphics[width=0.49\textwidth,keepaspectratio]{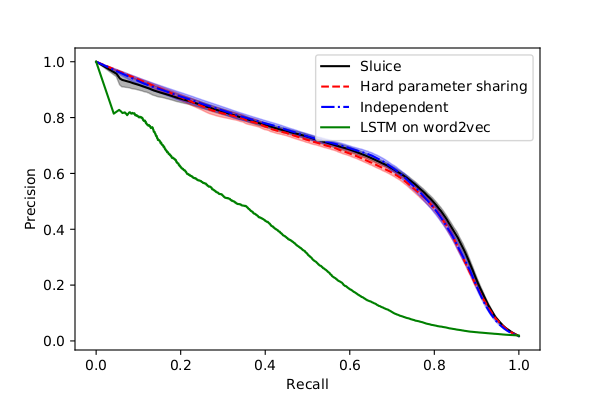}}
		\subfigure[ROC curves (log-scale for FPR)]{\includegraphics[width=0.49\textwidth,keepaspectratio]{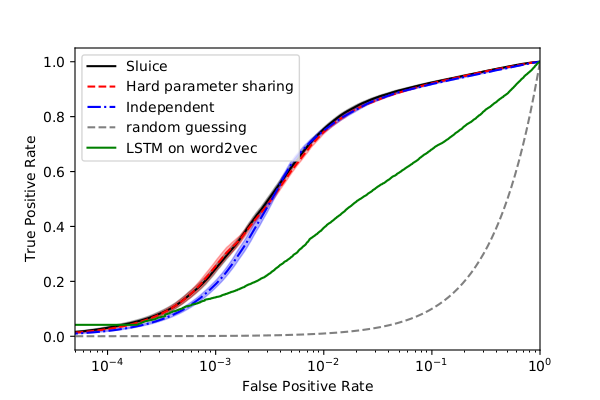}}
		\caption{Detection of infected clients.}
	\label{fig:future_data_comp}
\end{figure*}

\begin{figure*}[h!]
	\centering
		\subfigure[Precision-recall curves]{\includegraphics[width=0.49\textwidth,keepaspectratio]{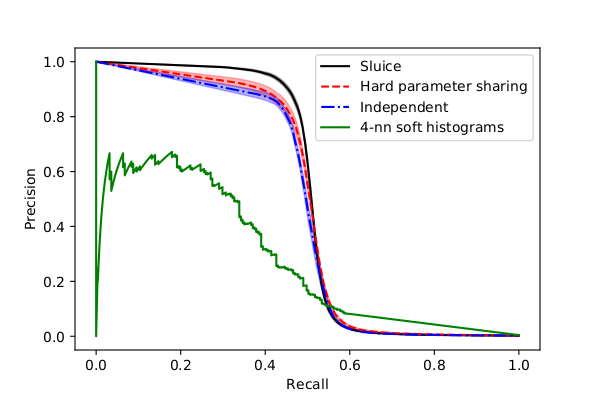}}
		\subfigure[ROC curves (log-scale for FPR)]{\includegraphics[width=0.49\textwidth,keepaspectratio]{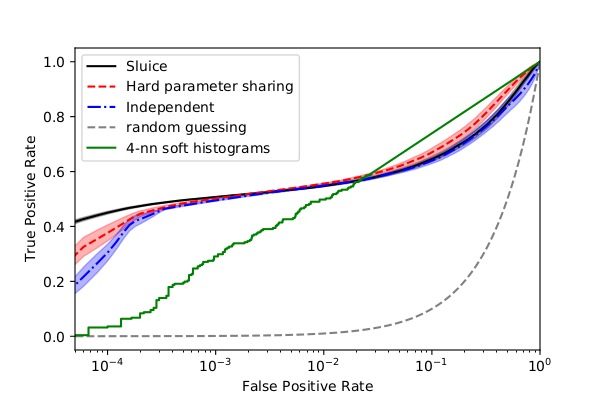}}
		\caption{Detection of malicious domains.}
	\label{fig:future_domain_comp}
\end{figure*}

\begin{figure*}[h!]
	\centering
		\subfigure[Infected clients: different malware families.]{\includegraphics[width=0.49\textwidth,keepaspectratio]{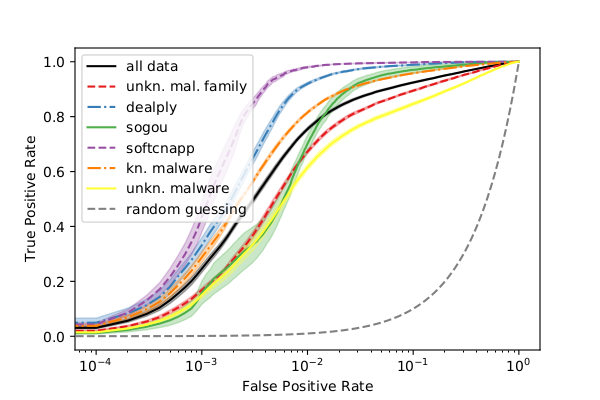}}
	  \subfigure[Infected clients: types of malware.]{\includegraphics[width=0.49\textwidth,keepaspectratio]{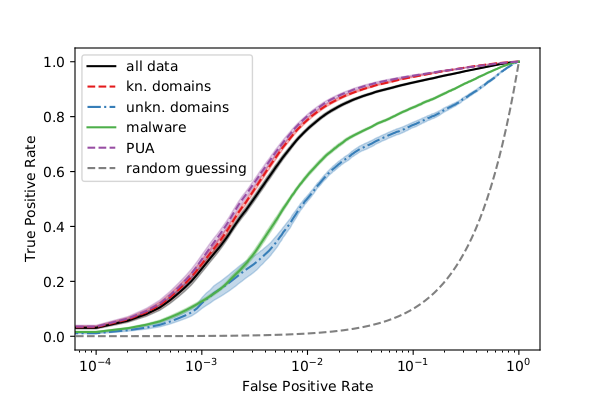}}
		\subfigure[Malicious domains: subgroups of domains.]{\includegraphics[width=0.49\textwidth,keepaspectratio]{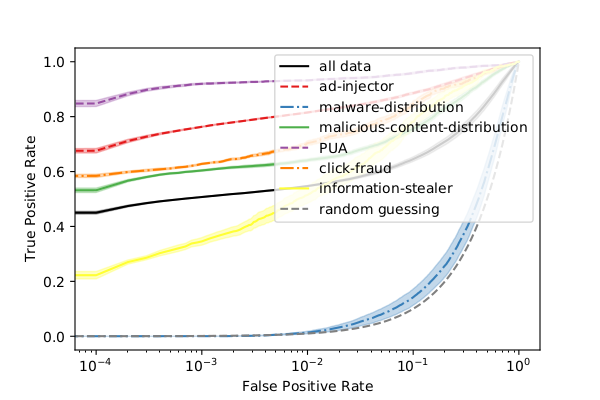}}
		\subfigure[Malicious domains: subgroups of domains.]{\includegraphics[width=0.49\textwidth,keepaspectratio]{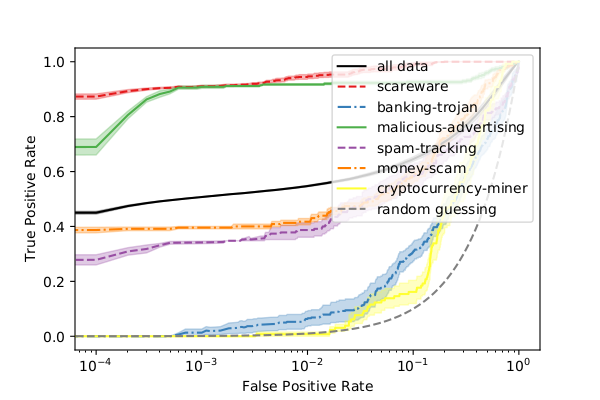}}
		\caption{{\sl Sluice network} on subgroups of instances, ROC curves for infected clients (left figure) and malicious domains (right figures).}\label{fig:detailed}
	\end{figure*}

\subsection{Infected Clients: Performance}

We train the models on the {\em training data} and evaluate them on {\em test data} that was recorded after the training data. 
Figure~\ref{fig:future_data_comp} compares precision-recall and ROC curves; curves are averaged over 10 random restarts with Glorot initialization, colored bands visualize plus and minus one standard error.

For malware detection, the {\sl sluice network}, the {\sl independent models} and {\sl hard parameter sharing} differ only marginally in performance. All three detect 40\% of malware with a precision of 80\%. Based on Welch's $t$-test with significance level $\alpha = 0.05$, at false-positive rates of $10^{-4}$ and $10^{-3}$, the {\sl sluice network} is still significantly better than the {\sl independent model} ($p=0.021$ for $10^{-4}$ and $p=0.008$ for $10^{-3}$), but the difference between {\sl sluice} and {\sl hard parameter sharing} is not significant. 
{\sl LSTM over word2vec} clearly performs substantially worse. 

\subsection{Malicious Domains: Performance}

Figure~\ref{fig:future_domain_comp} compares the model's performance for detection of malicious domains. Here, the precision-recall and ROC curves of the {\sl sluice network} look favorable compared to the baselines. Intuitively, since there are fewer malicious domains in the training data than there are infected clients, it is plausible that malicious-domain detection benefits more strongly from transfer learning. The {\sl 4-NN soft histogram} baseline performs substantially worse. The precision-recall curve becomes noisy near a recall of zero because for a low recall, the precision estimate is based on a small number of positives, and the decision function assumes a high value for several true and false positives.

Based on Welch's $t$-test with significance level $\alpha = 0.05$, at a false-positive rate of $10^{-4}$, the {\sl sluice network} performs significantly better than both the {\sl independent model} ($p=0.028$) and {\sl hard parameter sharing}. At $10^{-3}$, the {\sl sluice network} outperforms the {\sl independent model} ($p=0.001$); it detects 40\% of all malicious domains almost without false alarms. The difference between {\sl sluice} and {\sl hard parameter sharing} is not significant.

\subsection{Detailed Analysis}

In this section, we study how the detection models perform on specific subgroups of clients and domains. We train a single model on all training data.
In order to determine the performance for specific types of instances, we skip all other positive instances in the evaluation data. Since the class ratios vary widely between subgroups, we compare ROC curves. 
Figure~\ref{fig:detailed}(a) shows that the most popular malware families can be detected more easily, which corresponds to their high prevalence in the training data. Perhaps surprisingly, we see that the model detects unknown variations of known malware families as well as {\em unknown malware families}---that is, malware families for which no representative is present in the training data---almost as accurately as known malware. Figure~\ref{fig:detailed}(b) shows that the model's performance is just slightly better for the highly prevalent potentially unwanted applications (``PUA'') than it is for malware. 
We also see that malware which does not contact any domain that occurs in the training data (labeled ``unknown domains'') is detected with comparable accuracy to malware that contacts known domains. 

Figures~\ref{fig:detailed}(c and d) show how the {\sl sluice network} performs on specific types of malicious domains. Here, we see that the detection performance uniformly depends on the prevalence of the domain type in the training data. Only 10 backends for banking trojans are included in the training data, and no single cryptocurrency-mining backend. Malware-distribution servers are almost impossible for the model to detect, despite being the second-most frequent type of malicious domains in the training data. But a detailed analysis shows that the training data contains only 1,447 flows (out of more than 44 million) from malware-distribution servers; so at the level of flows, this class is actually rare.

\subsection{Additional Experiments}

We carry out additional experiments but omit the detailed results for brevity. The {\sl independent models} differ from {\sl LSTM on word2vec} in two aspects: the use of the domain-name CNN instead of a word2vec embedding, and the choice of processing windows of network flows in a convolutional way with max-pooling over all window positions instead of an LSTM. In order to explore whether performance differences are due to the different domain-name embedding or the different handling of the time-series input as additionally experiment with the intermediate form of {\em dense on word2vec}. We find that while this architecture performs significantly better than {\sl LSTM on word2vec}, it still performs much worse than {\sl independent models}. 

The {\sl 4-NN soft histogram model}~\citep{lokovc2016k} originally does not use the engineered domain features~\citep{franc2015learning} that we provide it with. We find that using the {\sl 4-NN} model without the domain features (or using the domain features without the histogram features) deteriorates the results. We also find that combining the soft-histogram features and engineered domain features with a random forest improves the result over the 4-NN classifier, but its performance remains substantially below the performance of all neural networks.
Finally, we find that adding additional convolutional and max-pooling layers or replacing the dense layer in the sluice network with convolutional and max-pooling layers deteriorates its performance.

\section{Conclusion}\label{sec:conclusion}

Detection of malware-infected clients and malicious domains allows organizations to use a centralized security solution that establishes a uniform security level across the organization with minimal administrative overhead. 
A specifically prepared VPN client makes it possible to collect large amounts of HTTPS network traffic and label network flows in retrospect by whether they originate from malware. This makes it possible to employ relatively high-capacity prediction models. By contrast, malicious domains have to be identified by means of an expensive forensic analysis. We have developed a method that jointly detects infected clients and malicious domains from encrypted network traffic without compromising the encryption.

We can draw a number of conclusions. 
All network architectures that we study improve on the previous state of the art by a large margin. 
We find that transfer learning using a sluice network improves malware-detection---for which we have a large body of training data---slightly over learning independent networks. Transfer learning allows us to leverage the large body of malware training data to improve the detection of malicious domains. The sluice network detects 40\% of all malware with a precision of 80\% using only encrypted HTTPS network traffic---at this threshold level, 20\% of all alarms are false alarms. In practice, each alarm triggers the notification of a security analyst; if 80\% of the notifications indicate an actual security breach, an analyst will not get the impression that the notification system can be ignored. The sluice network detects new variants of known malware families and malware of families that have not yet been known at training time with nearly the same accuracy. This finding is remarkable because signature-based antivirus tools cannot detect such malware. The network also detects 40\% of all malicious domains with a precision of nearly 1. Given the high costs of a manual analysis of domains, this result has a potentially high impact for network security in practice. 

\section*{Acknowledgment}
The work of Tom\'{a}\v{s} Pevn\'{y} has been partially funded by Czech Ministry of education under the GACR project 18-21409S. We would like to thank Virustotal.com for their kind support.

\bibliographystyle{plainnat}
\end{document}